\documentclass[letterpaper, 10 pt, conference]{ieeeconf}  

\IEEEoverridecommandlockouts                              

\usepackage{graphics} 
\usepackage{epsfig} 
\usepackage{times} 
\usepackage{amsfonts}
\usepackage{amssymb}  

\usepackage{graphicx}
\usepackage{amsmath}
\usepackage{algorithm}
\usepackage{algorithmic}
\usepackage{multirow}
\usepackage{soul}
\usepackage{color}
\usepackage{pifont}
\usepackage{pgfplots}
\usepackage[outline]{contour}
\usepackage{subcaption}
\usepackage{booktabs}
\usepackage{eqparbox}
\usepackage{pdfpages}

\usepackage[flushleft]{threeparttable} 

\usepgfplotslibrary{fillbetween}
\pgfplotsset{compat=1.10}
\usepackage{stfloats}
\definecolor{cvprblue}{rgb}{0.21,0.49,0.74}
\usepackage[pagebackref,breaklinks,colorlinks,citecolor=cvprblue]{hyperref}

\DeclareMathOperator*{\argmin}{arg\,min}

\usepackage{xspace}
\makeatletter
\DeclareRobustCommand\onedot{\futurelet\@let@token\@onedot}
\def\@onedot{\ifx\@let@token.\else.\null\fi\xspace}

\def\eg{\emph{e.g}\onedot} 
\def\ie{\emph{i.e}\onedot} 
 
\def\etc{\emph{etc}\onedot} 
 
\def\etal{\emph{et al}\onedot}
\makeatother

\title{\LARGE \bf Data selection method for assessment of autonomous vehicles}

\author{Linh Trinh \textit{Student Member IEEE}, Ali Anwar \textit{Member IEEE}, Siegfried Mercelis\\
{Faculty of Applied Engineering, IDLab, University of Antwerp-imec, Belgium} \\
{\tt\small\{linh.trinh, ali.anwar, siegfried.mercelis\}@uantwerpen.be}\\
}

\begin{document}

\maketitle
\thispagestyle{empty}
\pagestyle{empty}

\begin{abstract}
As the popularity of autonomous vehicles has grown, many standards and regulators, such as ISO, NHTSA, and Euro NCAP, require safety validation to ensure a sufficient level of safety before deploying them in the real world. Manufacturers gather a large amount of public road data for this purpose. However, the majority of these validation activities are done manually by humans. Furthermore, the data used to validate each driving feature may differ. As a result, it is essential to have an efficient data selection method that can be used flexibly and dynamically for verification and validation while also accelerating the validation process. In this paper, we present a data selection method that is practical, flexible, and efficient for assessment of autonomous vehicles. Our idea is to optimize the similarity between the metadata distribution of the selected data and a predefined metadata distribution that is expected for validation. Our experiments on the large dataset BDD100K show that our method can perform data selection tasks efficiently. These results demonstrate that our methods are highly reliable and can be used to select appropriate data for the validation of various safety functions. 
\end{abstract}

\section{INTRODUCTION}
Autonomous vehicles today are not only confined to the research labs but they are becoming more prevalent in the real-world. Recently, many automotive Original Equipment Manufacturers (OEMs) launched their commercial autonomous vehicles, such as Tesla, Alphabet, Waymo, and so on \cite{validation_odd,scenario_pdf,euro_ncap_2023}. To standardize autonomous vehicles worldwide, self-driving is usually divided into 6 levels, from 0 (zero self-driving) to 5 (full self-driving) \cite{odd@sae} where each level consists of a set of Advanced Driver-Assistance System (ADAS) or Automated Driving System (ADS) features such as Adaptive Cruise Control (ACC), Lane Keeping Assist (LKA), Highway Pilot (HWP), and others \cite{nhtsa_fw,odd@sae}. Following many standards such as ISO 26262, 21448, ANSI/UL 4600 \cite{iso2018_26262,iso2020tr,iso21448,ansi_ul_4600}, ISO 15622 \cite{iso_15622_2018}, ISO 17361 \cite{iso_17361_2017}, ISO 11270 \cite{iso_11270_2014}, ISO 19237 \cite{iso_19237_2017}, ISO 22839 \cite{iso_22839_2013}, and national regulators such as the USA NHTSA, EURO NCAP \cite{nhtsa_fw,euro_ncap_2018}, ASEAN NCAP \cite{asean_ncap}, autonomous vehicles must pass certain test cases and scenarios on the public road before being granted the permission to drive in the real world. Data is crucial in the functioning of autonomous vehicles as they heavily depend on vast quantities of it for validation purpose and to successfully implement autonomous vehicles in real-world scenarios. Since self-driving is a safety-critical application, validation requires a diverse set of testing scenarios that are representative of driving. Many OEMs, such as Alphabet, Tesla, and Audi, \etc, have collected a large amount of public road data \cite{validation_odd} for validating safety of their high-level autonomous vehicles \cite{scenario_pdf}. But collecting data on the public road can generate a huge amount of data on a daily basis. Processing all this data for validation purposes is an extremely exhausting and impractical effort. Hence it is more important to query a subset of the data which is smaller than the original dataset to accelerate validation process. 
The Society of Automotive Engineers (SAE) has established the Operational Design Domain (ODD) as a framework for determining the scope of validation test cases and scenarios conducted on public roads.
ODD then becomes a fundamental condition when validating an ADAS/ADS system of autonomous vehicles \cite{iso2020tr,iso21448,ansi_ul_4600,nhtsa_fw}. 
Several recent works attempted to define scenario identification in terms of corresponding ODD to pass validation, such as \cite{scenario_pdf,validation_odd}. However, because public road validation is highly manual, it is heavily reliant on human effort \cite{sae_curation@Perrin}. Moreover, the large amount of data presents a significant challenge for self-driving system validation.
Hence, locating and selecting a smaller set that covers the expected ODD is an effective way to reduce human effort while increasing productivity and efficiency. Furthermore, data selection logically implies data reduction. To save storage space, data may need to be discarded after collection or application completion. However, because data collection on public roads is prohibitively expensive, the data should be separated into useful information that can be used later. Mining these examples can be based on a variety of criteria, including unseen conditions, objects or scenes, or examples where self-driving systems underperformed.

Hence, a flexible and dynamic data selection approach is required for validation of ADAS/ADS features of autonomous vehicle. However, to the best of our knowledge, there is a lack of research or algorithm development pertaining to the aforementioned objectives. Recently, several commercial services, such as ADaaS\footnote{https://www.aimmo.ai/p/adaas}, Scale Nucleus\footnote{https://scale.com/nucleus}, dSPACE IVS\footnote{https://intempora.com/products/intempora-validation-suite}, and many others provide data curation services on data collected on the road. These services, on the other hand, primarily facilitate data querying based on user-defined filters which are relevant to ODD list. However, these commercial services are disclosed, with no explicit method or algorithm described. In this paper, we propose a data selection method for selecting subset of the data to validate multiple self-driving features. Our main idea is to perform the selection using the dataset's metadata. In more detail, we optimize the similarity of selected data's metadata distribution to the preferred metadata distribution of each validation task. The selection quality can then be evaluated using our proposed metrics and metadata distributions.

In summary, our main contributions are described below: 
\begin{itemize}
    \item We propose a data selection method for selecting a subset of data from a large data set for validating autonomous vehicles. Our method consists of a framework for data selection, an algorithm for training scoring model, which is the heart of our framework, and an algorithm for selecting data. 
    \item We propose two metrics to guide the selection algorithm as well as measure the selection quality.
    \item We conduct extensive experiments to evaluate and analyze our method. The results of selection tasks on a large public dataset (BDD100K video) containing over a thousand hours of data, demonstrate the performance of our method. 
\end{itemize}
In the following section, we discuss related works in Section \ref{sec:related_work}, then present technical details of our methodology in Section \ref{sec:method}, and show the results of our experiment in Section \ref{sec:exp}. Finally, Section \ref{sec:conclude} concludes our work.

\section{RELATED WORKS} \label{sec:related_work}
In reality, self-driving systems are required to validate not only the correctness of their system, but also their safety. Many standards such as ISO 26262, 21448, ANSI/UL 4600 \cite{iso2018_26262,iso2020tr,iso21448,ansi_ul_4600}, ISO 15622 \cite{iso_15622_2018}, ISO 17361 \cite{iso_17361_2017}, ISO 11270 \cite{iso_11270_2014}, ISO 19237 \cite{iso_19237_2017}, ISO 22839 \cite{iso_22839_2013} or national regulators such as NHTSA \cite{nhtsa_fw}, Euro NCAP \cite{euro_ncap_2018,euro_ncap_2023}, ASEAN NCAP \cite{asean_ncap} require ADAS/ADS system of autonomous vehicles to pass the validation on the public road for any SAE self-driving levels \cite{odd@sae}.
These standards and regulations provide the specification of test cases for autonomous vehicle safety validation, which typically consists of certain stages: data collection, event extraction (\eg scenario, tagging, \etc), data selection for each test case, and verification or validation. In recent years, many OEMs have focused on collecting public road data for validation. For example, Waymo has accumulated 5 million miles since 2018, Cruise collected over 770,000 miles in 2020 \cite{pp4av,fisheyepp4av}, BMW has been collecting more than 230 PB data from more than 100 vehicles since 2019 \cite{sae_curation@Perrin}. 
As stated by SAE \cite{odd@sae}, the ODD refers to the specific set of operating conditions that a particular driving automation system or its feature is designed to function within. This includes various factors such as environmental conditions, geographical limitations, time-of-day restrictions, and specific traffic or roadway characteristics.
In simplicity, ODD comprises of a collection of high-level domains as well as details on static tags and dynamic events known as scenarios. Many prior works concentrated on building algorithms for extracting scenarios from captured data, such as clustering-based scenario extraction \cite{rel_cluster_scenario}, corner-cases scenario finding \cite{rel_corner_case}, traffic scenario extraction \cite{rel_scenario_extract}, tags extraction \cite{rel_universal_tag}. In addition to these techniques, self-driving fleets can perform scenarios manually by triggering specified events during test driving \cite{sae_curation@Perrin}. In the following stage, data for each scenario needs to be curated for validation or verification. Data curation is typically accomplished using metadata tags or scenarios that have already been retrieved and saved alongside the acquired data, as defined by ODD \cite{sae_curation@Perrin}. There are two types of data selection: frame-level and scenario-level. The majority of selection processes are performed manually by humans, which is related with the manual validation process \cite{sae_curation@Perrin}. Due to the enormous volume of collected data and the limited availability of validation resources, the output of data selection is typically smaller than the original dataset. Furthermore, with a high number of test cases, the selection operation must run often. A few works describe methods for selecting scenarios for a specific test case, such as \cite{rel_select_scenario,rel_iden_scenario}. Sadat \etal \cite{selection_scenario@Sadat} describe a dataset selection approach that uses infrastructure, traffic participants, and driving maneuvers to measure the level of interest in traffic scenes. The goal of this method is to improve the performance of deep learning models used in self-driving perception tasks. Bogdoll \etal \cite{selection_corner_case@Bogdoll} proposed an algorithm for selecting corner case scenarios.
Unfortunately, to the best our knowledge, there is a lack of scientific study on data selection to support the validation of autonomous vehicles. Several commercial services, such as the above-mentioned ADaaS, Scale Nucleus, dSPACE IVS, and so on, provide data curation services on data collected on the road for fleet testing or holomogation. These services, on the other hand, primarily facilitate data querying based on user-defined filters such as ODD condition filtering. Moreover, these methods do not provide an explicit algorithm or methodology inside. We then propose a method to better support flexible and dynamic selection for validating autonomous vehicles in the next section.

\section{METHOD} \label{sec:method}
This section describes the details of our method. Our primary idea for scenario selection for autonomous vehicle verification and validation is to use the dataset's metadata. Metadata can be broadly defined as the compilation of information from multiple sources in a dataset. For example, the dataset includes environmental data like weather and road types, as well as dynamic traffic participants like cars, buses, trucks, and pedestrians. Similar to ODD, we consider metadata to be constructed from domains, each of which might contain a set of distinct categories (or tags). Several domains of metadata can coexist independently. For example, data collected on public roads may include weather information in addition to road information; the weather domain is divided into four categories: wet, sunny, foggy, and cloudy. Similarly, there are five domain object categories: vehicle, bus, truck, pedestrian, and motorcycle. 
 We assume that in the early stages of any data-driven validation task, experts have an expectation of a metadata distribution of interest.
 For instance, when choosing a subset of data to verify a lane assistant keeping feature, the priority is given to selecting roads with lanes rather than dirt or gravel roads. Therefore, the goal is to ensure that the selected data contains lane-roads. The predetermined expectation of metadata distribution is denoted as $\mathsf{E}$, which $\mathsf{E}\in \mathbb{R}^{D \times C}$ where $D$, and $C$ denote the number of distinct domains, and the number of categories inside each domain, respectively. Table \ref{table:ex_metadata} shows an illustrative instance of expectation, featuring two domains and their corresponding categories.
\begin{table}[ht]
\centering
\caption{Example of a metadata distribution.}
\footnotesize{
\begin{tabular}{|c|c|c|c|c|c|}
\hline
\textbf{Domain}            & \textbf{Category} & \textbf{Ratio} & \textbf{Domain}          & \textbf{Category} & \textbf{Ratio} \\ \hline
\multirow{4}{*}{Road type} & Highway           & 0.6             & \multirow{4}{*}{Weather} & Sunny             & 0.8             \\ \cline{2-3} \cline{5-6} 
                           & Urban             & 0.2             &                          & Rainy             & 0.17             \\ \cline{2-3} \cline{5-6} 
                           & Rural             & 0.1             &                          & Foggy             & 0.03              \\ \cline{2-3} \cline{5-6} 
                           & Other             & 0.1             &                          & Snowy             & 0.0              \\ \hline
\end{tabular}
}
\label{table:ex_metadata}
\end{table}
\begin{figure*}[ht]
    \centering
    \includegraphics[width=0.98\textwidth]{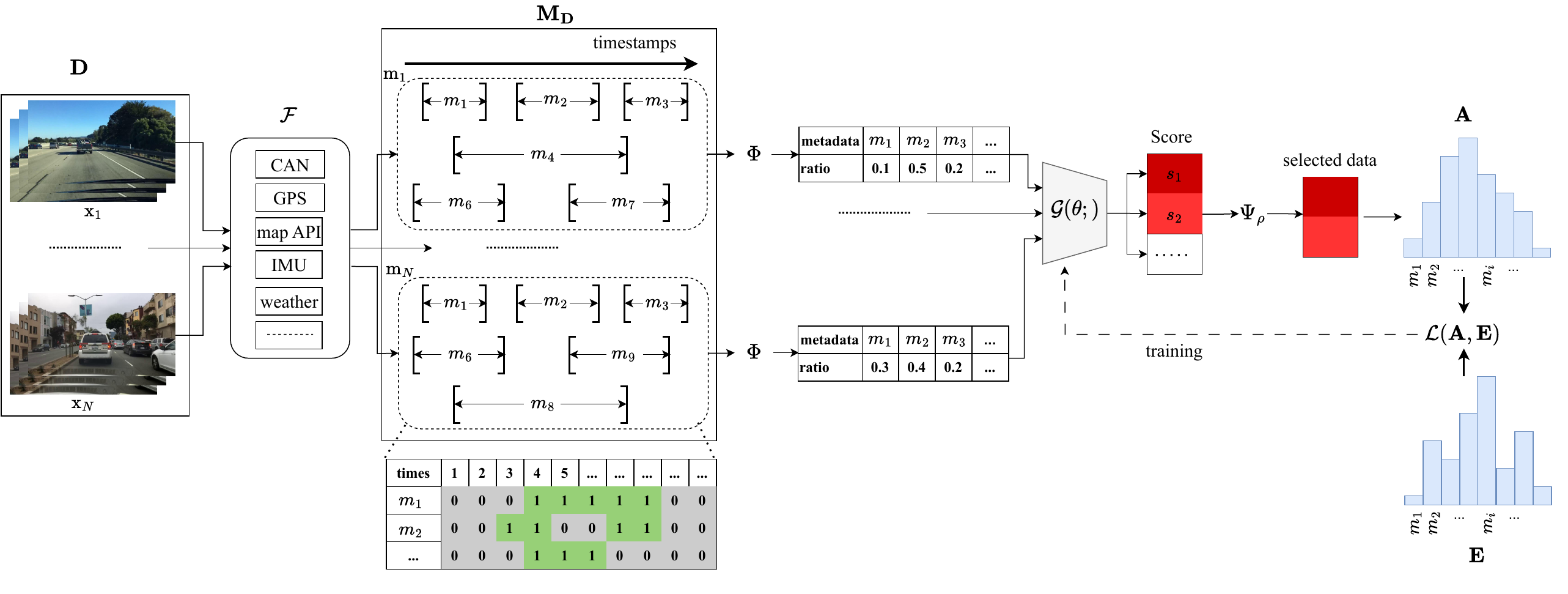}
    \caption{Illustration of data selection workflow in our proposed framework.}
    \label{fig:fw}
\end{figure*}

Denote the original dataset $\mathbf{D}=\{\mathrm{x}_i \}_{i=1}^N$ with $N$ data points, and $|\cdot|$ represents the dataset size. Each data point $\mathrm{x}_i$ can be a scenario data, which may include a sequence of frames, GPS data, CAN signal, etc. The goal is to find a subset dataset $\mathbf{U}\subseteq \mathbf{D}$ that is smaller than $\mathbf{D}$. We select a smaller dataset with a given ratio $\rho=\frac{|\mathbf{U}|}{|\mathbf{D}|}$, where $\rho$ is the keep ratio and $0\leq \rho \leq 1$. Our main idea is to maximize the distribution of metadata for the selected data $\mathbf{U}$ so that it matches the expected metadata distribution $\mathsf{E}$. Let $\mathsf{A}$ denote the metadata distribution of the selected data $\mathbf{U}$, where $\mathsf{A}\in \mathbb{R}^{D \times C}$.
We define two metrics which are category-based metric $\mathcal{S}_c$ and domain-based metric $\mathcal{S}_d$, referred to as equation \ref{eq:score_regular} and equation \ref{eq:score_scale}, to measure the similarity between two metadata distributions.
\begin{equation} \label{eq:score_regular}
    \mathcal{S}_c(\mathsf{A}, \mathsf{E}) = 1- \frac{\lVert \mathsf{A} - \mathsf{E} \rVert }{D}
\end{equation}
\begin{equation}\label{eq:score_scale}
    \mathcal{S}_{d}(\mathsf{A}, \mathsf{E}) = 1- \frac{\sum_{i=1}^{|\mathsf{A}|} \mathbf{d}_i }{|\mathsf{A}|}
\end{equation}
where $\mathrm{d}_i$ is defined as in the equation \ref{eq:score_item}. 
\begin{equation} \label{eq:score_item}
    \mathrm{d}_i = \zeta(\frac{\lVert\mathsf{A}_i - \mathsf{E}_i\rVert}{\mathsf{E}_i})
\end{equation}
with $\zeta(\alpha)$ is $\min(1,\alpha)$ for simplicity.
While the domain-based metric indicates how good the selection method is when compared to the average of domains, the category-based metric indicates how good the selection method is when compared to the average of categories.
These metrics can be used to measure the quality of selection directly when the selection was done. If both of these metrics are higher then it indicates a better match with expected $\mathsf{E}$.

The framework in Figure \ref{fig:fw} illustrated the workflow of our data selection. The metadata can be obtained by using the metadata extraction module $\mathcal{F}$, which can be a set of functions such as map API to obtain geographic information, car information (\eg acceleration, velocity, throttle info) from CAN, weather information query, and so on. The output of metadata extraction $\mathcal{F}$ for each sample $\mathrm{x}_i$ is the series of metadata $\mathrm{m}_i \in \mathbf{M}_{\mathbf{D}}$ associated with this sample $\mathrm{x}_i$, where $\mathbf{M}_{\mathbf{D}}$ denotes all extracted metadata of the entire data $\mathbf{D}$. Each $\mathrm{m}_i$ contains $M$ interested metadata tags, $m_1,m_2,...,m_M$. Metadata $\mathrm{m}_i$ can be a series of metadata tag by time, with the value associated with each metadata tag $m_j$ set to 1 if it occurred and 0 otherwise. Each extracted metadata tag $m_j$ by $\mathcal{F}$ can be represented as a series in time, for example, a highway tag of data occurred every 5 seconds. To summarize the extracted metadata $\mathrm{m}_i$ of each sample $\mathrm{x}_i$, we use a transformation function $\Phi$ to transform metadata to a ratio list. The ratio of each metadata tag $m_j$ can be the duration of this metadata tag over the duration of the data sample $\mathrm{x}_i$.
All metadata distribution $\{\Phi(\mathrm{m}_i)|i=1,..,N\}$ will be inputed into a scoring model $\mathcal{G}(\theta;)$ for calculating the important score of data $\mathrm{x}_i$, which is later used for selection decision, where $\theta$ is the parameter of model $\mathcal{G}$. The calculated score will be used by a selection function $\Psi_{\rho}$, which seeks to keep a rate $\rho$ of data. To simplify, we use the function $\Psi_{\rho}$ as the selection function, which selects the top $\rho$ items from the ranking of $\mathbf{D}$. The ranking can be used to determine the level of significance or interest, as well as diversity and complexity. Thus, we select $\Psi$ as the descending rank in the scoring function $\mathcal{G}$. For example, when comparing two data points, $\mathrm{x}_1$ and $\mathrm{x}_2$, the data point with the higher score is more likely to be retained. Generally, if $\mathcal{G}(\theta,\Phi(\mathrm{m}_h)) > \mathcal{G}(\theta,\Phi(\mathrm{m}_k))$, then $\mathrm{x}_h$ is more likely to be retained than $\mathrm{x}_k$.

A loss function $\mathcal{L}(\mathsf{A},\mathsf{E})$ will be conducted from the actual distribution $\mathsf{A}$ of selected data and a predefined expected metadata distribution $\mathsf{E}$ to train model $\mathcal{G}(\theta;)$. We formulate the problem of selection as equation \ref{eq:opt}.
\begin{equation} \label{eq:opt}
    \begin{aligned}
        \min_{\theta} \quad & \mathcal{L}\left(\mathcal{G}(\theta;\mathbf{M}_{\mathbf{D}}), \mathsf{E}\right)\\
        \textrm{s.t.} \quad & \frac{|\Psi(\mathbf{M}_{\mathbf{D}})|}{|\mathbf{M}_{\mathbf{D}}|}=\rho\\
    \end{aligned}
\end{equation}
The selection problem now becomes to solve the optimization problem below for finding optimal model $\mathcal{G}$:
\begin{equation}\label{eq:optimization_prob}
    \operatorname*{\argmin}_\theta \mathcal{L}\left(\mathcal{G}(\theta;\mathbf{M}_{\mathbf{D}}), \mathsf{E}\right)
\end{equation}

\begin{algorithm}[ht]
\caption{Optimizer for finding optimal $\mathcal{G}(\theta^*;)$}
\label{al:solver}
    \hspace*{\algorithmicindent} \textbf{Input:} metadata $\mathbf{M}_{\mathbf{D}}$, expected distribution $\mathsf{E}$  \\
    \hspace*{\algorithmicindent} \textbf{Output:} optimal $\mathcal{G}(\theta^*;)$ 
    \begin{algorithmic}[1] 
        \STATE Initialized number of epoch $T$, batch size $K$
        \STATE Initialize parameter $\theta$
        \STATE Calculate the number of batches $B=\lceil \frac{|\mathbf{M}_{\mathbf{D}}|}{K} \rceil$
        \FOR{\text{all} $t =1, ...\text{ }..., T$}
            \WHILE{$b\leq B$}
                \STATE \text{Sample batch} $\mathbf{M}_\mathbf{D}^b$ \text{from metadata}
                \STATE \text{Forward:} \textit{calculate score} $\mathcal{G}(\theta;\Phi(\mathbf{M}_\mathbf{D}^b))$
                \STATE $\mathbf{M}_{sub} \leftarrow \Psi_\rho(\mathcal{G}(\theta;\Phi(\mathbf{M}_\mathbf{D}^b)))$
                \STATE $\mathsf{A} \leftarrow \Phi(\mathbf{M}_{sub})$
                \STATE \text{Calculate loss:} $\mathcal{L} \leftarrow 1-\mathcal{S}_{c|d}(\mathsf{A}, \mathsf{E})$
                \STATE \text{Compute gradient: } $\mathcal{L}.\textit{backward()}$
                \STATE \text{Update parameters} $\theta$: \textit{optimizer.step()}
             \ENDWHILE
        \ENDFOR
        \STATE \textbf{return} Optimal $\mathcal{G}(\theta^*;)$
    \end{algorithmic}
\end{algorithm}
We propose the algorithm \ref{al:solver} for optimizing model $\mathcal{G}$. We set up model $\mathcal{G}$ as a neural network, with its parameter (\ie $\theta$) representing the weight of the neural network. 

\begin{algorithm}[ht]
\caption{Selected function $\Psi_\rho$}
\label{al:select}
    \hspace*{\algorithmicindent} \textbf{Input:} $\mathbf{M}_{\mathbf{D}}=\{\mathrm{m}_1,...,\mathrm{m}_n\}$, score $S=\{s_1,...,s_n\}$  \\
    \hspace*{\algorithmicindent} \textbf{Output:} metadata $\mathbf{M}_{sub}$
    \begin{algorithmic}[1] 
        \STATE Initialized number selected data $N_s\leftarrow \lceil\rho*n\rceil$
        \STATE Initialize $\mathbf{M}_{sub} \leftarrow \{\}$
        \WHILE{$|\mathbf{M}_{sub}| < N_s$}
            \STATE $S_{tmp} = S$
            \WHILE{$S_{tmp} \neq \emptyset$}
                \STATE $i \leftarrow argmax (S_{tmp})$ ; $S_{tmp}\leftarrow S_{tmp}\setminus\{s_i\}$
                \STATE \textit{check} $\mathcal{D}(\mathrm{m}_i, \mathrm{m}_j) < \epsilon, \forall \mathrm{m}_j\in \mathbf{M}_{sub}$
                \STATE $\mathbf{M}_{sub} \leftarrow \mathbf{M}_{sub} \cup \{\mathrm{m}_i\}$; $S\leftarrow S\setminus\{s_i\}$
            \ENDWHILE
            \STATE $\epsilon \leftarrow \epsilon\times\eta$
         \ENDWHILE
        \STATE \textbf{return} metadata $\mathbf{M}_{sub}$
    \end{algorithmic}
\end{algorithm}
To increase the diversity of selected data, we use data similarity filtering in the selected function $\Psi_\rho$. Our idea is to avoid keeping data that is similar to what has already been selected. The algorithm \ref{al:select} describes the details of the selected $\Psi_\rho$ process. There are several idea to measure the similarity of two data \cite{multipledatasourcesdomain,seadsc}, but we use the cosine similarity function which is commonly used for similarity measurement, to determine the similarity of two metadata vectors. So, we have the similarity score of two metadata $\mathrm{m}_i$ and $\mathrm{m}_j$ as below:
\begin{equation}
    \mathcal{D}(\mathrm{m}_i,\mathrm{m}_j) = \mathcal{D}(\Phi(\mathrm{m}_i),\Phi(\mathrm{m}_j))=\frac{\vec{\Phi(\mathrm{m}_i)}\cdot\vec{\Phi(\mathrm{m}_j)}}{|\Phi(\mathrm{m}_i)||\Phi(\mathrm{m}_j)|}
\end{equation}
The high value of the threshold $\epsilon$ may cause the selected function $\Psi_\rho$ to not reach the expected selection rate $\rho$. To guarantee the outcome of selection while maintaining data similarity filtering, we adjust $\epsilon$ by a rate $\eta (0<\eta<1)$.

\textbf{Limitations.} The goal of our selection method is to optimize the representation of the selected dataset to match the predefined expected ratio. The expected ratio must be provided based on the human knowledge depending on the application. A poor predefined expected ratio may result in poor performance on downstream applications. 

\section{EXPERIMENT} \label{sec:exp}
In this section, we run extensive experiments to assess the effectiveness of our proposed method.
\subsection{Settings}
\textbf{Dataset.} We use BDD100K \cite{bdd100k} video, a very large open self-driving dataset which provided GPS signal for extracting metadata. This dataset contains 100,000 videos collected across the United States of America in the more than 300 days. Each video is about 40 seconds long and 30 frames per second. In this experiment, our goal is to select a subset of videos from the entire video dataset. The videos also include GPS and IMU data captured by cell phones. As the metadata query from OpenStreet map service \cite{openstreetmap} via GPS, we implement the metadata extraction function $\mathcal{F}_M$. We extract the 8 domains and their categories using GPS, as shown in Table \ref{table:ratio}. Because some GPS positions from cell phones cannot be queried in Openstreet map, therefore out of 100,000 videos, 72,197 videos with metadata totaling 2,861,030 seconds were successfully extracted.
The ratio between the duration of all the videos in the category with the total duration represented in all the categories is represented as the \textit{Original} column in the Table \ref{table:ratio}. 

\textbf{Our method.} For model training, we use a $\mathcal{G}$ model with two layers of 128 hidden layers each. The batch size $K$ is set at 1024. We train our model with the Adam optimizer and a learning rate of $0.01$. The model was trained for $T=120$ epochs. We set the initialized $\epsilon$ to $0.9$ and $\eta=0.85$.  We set the selection ratio $\rho$ in a wide range in several experiments. Due to lack of scientific methods and studies on selecting data for validating autonomous vehicles, we chose the work of selecting a diversity and complexity data set based on geographical information \cite{selection_scenario@Sadat} (DC). 

\subsection{Results}
\begin{table*}[ht]
\centering
\caption{Performance of selection on our method and DC \cite{data_curation} on various selection ratios $\rho$. ('-': skipping category, which is not included in the counting metric.)}
 \resizebox{\textwidth}{!}{%
 \begin{threeparttable}
\begin{tabular}{|ccc|ccc||ccc||ccc||ccc|}\hline
\multicolumn{2}{|c|}{Metadata}                                                             & \multicolumn{1}{c|}{\multirow{2}{*}{Original}} & \multicolumn{3}{c||}{$\rho=0.2$}                                                             & \multicolumn{3}{c||}{$\rho=0.4$\tnote{*}}                                                             & \multicolumn{3}{c||}{$\rho=0.6$\tnote{*}}                                                             & \multicolumn{3}{c|}{$\rho=0.8$\tnote{*}}                                                             \\ \cline{1-2} \cline{4-15} 
\multicolumn{1}{|c|}{Domain}                          & \multicolumn{1}{l|}{Category}      & \multicolumn{1}{c|}{}                           & \multicolumn{1}{l|}{$\mathsf{E}$}          & \multicolumn{1}{l|}{DC \cite{data_curation}}               & \textbf{Ours.}                      & \multicolumn{1}{l|}{$\mathsf{E}$}          & \multicolumn{1}{l|}{DC \cite{data_curation}}               & \textbf{Ours.}                      & \multicolumn{1}{l|}{$\mathsf{E}$}          & \multicolumn{1}{l|}{DC \cite{data_curation}}               & \textbf{Ours.}                      & \multicolumn{1}{l|}{$\mathsf{E}$}          & \multicolumn{1}{l|}{DC \cite{data_curation}}               & \textbf{Ours.}                      \\ \hline

\multicolumn{1}{|l|}{\multirow{4}{*}{Way type}}       & \multicolumn{1}{l|}{Highway}       & 0.096                      & \multicolumn{1}{l|}{\underline{0.05}}        & \multicolumn{1}{l|}{0.0795}            & \textbf{0.0534}            & \multicolumn{1}{l|}{\underline{0.15}}        & \multicolumn{1}{l|}{0.084}           & \textbf{0.1994} & \multicolumn{1}{l|}{\underline{0.25}}        & \multicolumn{1}{l|}{0.1284}          & \textbf{0.147}  & \multicolumn{1}{l|}{\underline{0.1}}         & \multicolumn{1}{l|}{0.2326}          & \textbf{0.102}  \\ \cline{2-15} 
\multicolumn{1}{|l|}{}                                & \multicolumn{1}{l|}{Primary way}   & 0.22779                    & \multicolumn{1}{l|}{\underline{0.15}}        & \multicolumn{1}{l|}{0.4567}            & \textbf{0.2032}            & \multicolumn{1}{l|}{\underline{0.25}}        & \multicolumn{1}{l|}{0.3123}          & \textbf{0.2846} & \multicolumn{1}{l|}{\underline{0.25}}        & \multicolumn{1}{l|}{0.3017}          & \textbf{0.2631} & \multicolumn{1}{l|}{\underline{0.3}}         & \multicolumn{1}{l|}{0.2754}          & \textbf{0.2842} \\ \cline{2-15} 
\multicolumn{1}{|l|}{}                                & \multicolumn{1}{l|}{Secondary way} & 0.23083                    & \multicolumn{1}{l|}{\underline{0.1}}         & \multicolumn{1}{l|}{0.4227}            & \textbf{0.2109}            & \multicolumn{1}{l|}{\underline{0.25}}        & \multicolumn{1}{l|}{0.2102}          & \textbf{0.2835} & \multicolumn{1}{l|}{\underline{0.25}}        & \multicolumn{1}{l|}{0.2767}          & \textbf{0.2558} & \multicolumn{1}{l|}{\underline{0.3}}         & \multicolumn{1}{l|}{0.2709}          & \textbf{0.287}  \\ \cline{2-15} 
\multicolumn{1}{|l|}{}                                & \multicolumn{1}{l|}{Link way}      & 0.04974                    & \multicolumn{1}{l|}{\underline{0.01}}        & \multicolumn{1}{l|}{0.0241}            & \textbf{0.0356}            & \multicolumn{1}{l|}{\underline{0.1}}         & \multicolumn{1}{l|}{0.0427}          & \textbf{0.1129} & \multicolumn{1}{l|}{\underline{0.15}}        & \multicolumn{1}{l|}{\textbf{0.0781}} & 0.0617          & \multicolumn{1}{l|}{\underline{0.05}}        & \multicolumn{1}{l|}{0.1168}          & \textbf{0.0562} \\ \hline
\multicolumn{1}{|l|}{\multirow{6}{*}{Number of lane}} & \multicolumn{1}{l|}{1-lane}        & 0.5327                     & \multicolumn{1}{l|}{\underline{0.7}}         & \multicolumn{1}{l|}{0.2576}            & \textbf{0.7549}            & \multicolumn{1}{l|}{\underline{0.17}}        & \multicolumn{1}{l|}{0.2245}          & \textbf{0.1935} & \multicolumn{1}{l|}{\underline{0.1}}         & \multicolumn{1}{l|}{0.1539}          & \textbf{0.1263} & \multicolumn{1}{l|}{\underline{0.2}}         & \multicolumn{1}{l|}{0.1428}          & \textbf{0.206}  \\ \cline{2-15} 
\multicolumn{1}{|l|}{}                                & \multicolumn{1}{l|}{2-lanes}       & 0.1408                     & \multicolumn{1}{l|}{\underline{0.15}}        & \multicolumn{1}{l|}{\textbf{0.1488}}   & 0.0092                     & \multicolumn{1}{l|}{\underline{0.27}}        & \multicolumn{1}{l|}{\textbf{0.2496}} & 0.3013          & \multicolumn{1}{l|}{\underline{0.25}}        & \multicolumn{1}{l|}{0.1726}          & \textbf{0.2065} & \multicolumn{1}{l|}{\underline{0.15}}        & \multicolumn{1}{l|}{0.1371}          & \textbf{0.159}  \\ \cline{2-15} 
\multicolumn{1}{|l|}{}                                & \multicolumn{1}{l|}{3-lanes}       & 0.1531                     & \multicolumn{1}{l|}{\underline{0.05}}        & \multicolumn{1}{l|}{0.2439}            & \textbf{0.01}              & \multicolumn{1}{l|}{\underline{0.17}}        & \multicolumn{1}{l|}{0.2233}          & \textbf{0.193}  & \multicolumn{1}{l|}{\underline{0.25}}        & \multicolumn{1}{l|}{0.1652}          & \textbf{0.2378} & \multicolumn{1}{l|}{\underline{0.15}}        & \multicolumn{1}{l|}{0.1779}          & \textbf{0.1592} \\ \cline{2-15} 
\multicolumn{1}{|l|}{}                                & \multicolumn{1}{l|}{4-lanes}       & 0.1139                     & \multicolumn{1}{l|}{\underline{0.05}}        & \multicolumn{1}{l|}{0.3091}            & \textbf{0.2233}            & \multicolumn{1}{l|}{\underline{0.17}}        & \multicolumn{1}{l|}{0.2043}          & \textbf{0.1564} & \multicolumn{1}{l|}{\underline{0.2}}         & \multicolumn{1}{l|}{0.1369}          & \textbf{0.1855} & \multicolumn{1}{l|}{\underline{0.1}}         & \multicolumn{1}{l|}{0.1204}          & \textbf{0.1101} \\ \cline{2-15} 
\multicolumn{1}{|l|}{}                                & \multicolumn{1}{l|}{5-lanes}       & 0.0306                     & \multicolumn{1}{l|}{-}                 & \multicolumn{1}{l|}{-}                 & -                          & \multicolumn{1}{l|}{\underline{0.05}}        & \multicolumn{1}{l|}{\textbf{0.0503}} & 0.051           & \multicolumn{1}{l|}{\underline{0.1}}         & \multicolumn{1}{l|}{0.0387}          & \textbf{0.0429} & \multicolumn{1}{l|}{\underline{0.05}}        & \multicolumn{1}{l|}{0.0316}          & \textbf{0.0365} \\ \cline{2-15} 
\multicolumn{1}{|l|}{}                                & \multicolumn{1}{l|}{6-lanes}       & 0.0152                     & \multicolumn{1}{l|}{-}                 & \multicolumn{1}{l|}{-}                 & -                          & \multicolumn{1}{l|}{\underline{0.05}}        & \multicolumn{1}{l|}{0.03}            & \textbf{0.035}  & \multicolumn{1}{l|}{\underline{0.1}}         & \multicolumn{1}{l|}{0.0219}          & \textbf{0.0234} & \multicolumn{1}{l|}{\underline{0.01}}        & \multicolumn{1}{l|}{0.0183}          & \textbf{0.011}  \\ \hline
\multicolumn{1}{|l|}{Bridge}                          & \multicolumn{1}{l|}{Bridge}        & 0.0508                     & \multicolumn{1}{l|}{\underline{0.01}}        & \multicolumn{1}{l|}{0.0535}            & \textbf{0.0283}            & \multicolumn{1}{l|}{\underline{0.4}}         & \multicolumn{1}{l|}{0.101}           & \textbf{0.1115} & \multicolumn{1}{l|}{\underline{0.3}}         & \multicolumn{1}{l|}{\textbf{0.0831}} & 0.0755          & \multicolumn{1}{l|}{\underline{0.05}}        & \multicolumn{1}{l|}{0.0633}          & \textbf{0.0602} \\ \hline
\multicolumn{1}{|l|}{One way}                         & \multicolumn{1}{l|}{One way}       & 0.6101                     & \multicolumn{1}{l|}{\underline{0.6}}         & \multicolumn{1}{l|}{0.8581}            & \textbf{0.7475}            & \multicolumn{1}{l|}{\underline{0.5}}         & \multicolumn{1}{l|}{0.6698}          & \textbf{0.4374} & \multicolumn{1}{l|}{\underline{0.7}}         & \multicolumn{1}{l|}{0.7674}          & \textbf{0.6977} & \multicolumn{1}{l|}{\underline{0.5}}         & \multicolumn{1}{l|}{0.6028}          & \textbf{0.5321} \\ \hline
\multicolumn{1}{|l|}{Toll}                            & \multicolumn{1}{l|}{Toll}          & 0.0112                     & \multicolumn{1}{l|}{\underline{0.01}}        & \multicolumn{1}{l|}{0.0351}            & \textbf{0.0056}            & \multicolumn{1}{l|}{\underline{0.2}}         & \multicolumn{1}{l|}{0.0274}          & \textbf{0.0278} & \multicolumn{1}{l|}{\underline{0.1}}         & \multicolumn{1}{l|}{\textbf{0.0162}} & 0.0169          & \multicolumn{1}{l|}{\underline{0.1}}         & \multicolumn{1}{l|}{\textbf{0.0113}} & 0.0134          \\ \hline
\multicolumn{1}{|l|}{Tunnel}                          & \multicolumn{1}{l|}{Tunnel}        & 0.0032                     & \multicolumn{1}{l|}{\underline{0.01}}        & \multicolumn{1}{l|}{0.0022}            & \textbf{0.0056}            & \multicolumn{1}{l|}{\underline{0.35}}        & \multicolumn{1}{l|}{0.002}           & \textbf{0.0035} & \multicolumn{1}{l|}{\underline{0.1}}         & \multicolumn{1}{l|}{0.0016}          & \textbf{0.0052} & \multicolumn{1}{l|}{\underline{0.05}}        & \multicolumn{1}{l|}{0.0012}          & \textbf{0.004}  \\ \hline
\multicolumn{1}{|l|}{Roundabout}                      & \multicolumn{1}{l|}{Roundabout}    & 0.001                      & \multicolumn{1}{l|}{\underline{0.01}}        & \multicolumn{1}{l|}{0.0024}            & \textbf{0.0034}            & \multicolumn{1}{l|}{\underline{0.35}}        & \multicolumn{1}{l|}{0.001}           & \textbf{0.0025} & \multicolumn{1}{l|}{\underline{0.05}}        & \multicolumn{1}{l|}{0.0009}          & \textbf{0.011}  & \multicolumn{1}{l|}{\underline{0.05}}        & \multicolumn{1}{l|}{0.0006}          & \textbf{0.0012} \\ \hline
\multicolumn{1}{|l|}{Shoulder}                        & \multicolumn{1}{l|}{Shoulder}      & 0.0001                     & \multicolumn{1}{l|}{-}                 & \multicolumn{1}{l|}{-}                 & -                          & \multicolumn{1}{l|}{\underline{0.5}}         & \multicolumn{1}{l|}{0}               & \textbf{0.0002} & \multicolumn{1}{l|}{\underline{0.01}}        & \multicolumn{1}{l|}{0}               & \textbf{0.0001} & \multicolumn{1}{l|}{\underline{0.05}}        & \multicolumn{1}{l|}{\textbf{0.0001}} & \textbf{0.0001} \\ \hline

\bottomrule

\multicolumn{3}{|c|}{\textbf{$\mathcal{S}_c$}}                                                                                    & \multicolumn{1}{c|}{\multirow{2}{*}{}} & \multicolumn{1}{c|}{0.7269} & \textbf{0.8887} & \multicolumn{1}{c|}{\multirow{2}{*}{}} & \multicolumn{1}{c|}{0.7217} & \textbf{0.7556} & \multicolumn{1}{c|}{\multirow{2}{*}{}} & \multicolumn{1}{c|}{0.8462} & \textbf{0.887} & \multicolumn{1}{c|}{\multirow{2}{*}{}} & \multicolumn{1}{c|}{0.9061}  & \textbf{0.9551} \\ \cline{1-3} \cline{5-6} \cline{8-9} \cline{11-12} \cline{14-15} 
\multicolumn{3}{|c|}{\textbf{$\mathcal{S}_d$}}                                                                                    & \multicolumn{1}{l|}{}                  & \multicolumn{1}{c|}{0.2154} & \textbf{0.4136}   & \multicolumn{1}{l|}{}                  & \multicolumn{1}{c|}{0.5246} & \textbf{0.6119} & \multicolumn{1}{c|}{}                  & \multicolumn{1}{c|}{0.4409} & \textbf{0.5331}  & \multicolumn{1}{c|}{}                  & \multicolumn{1}{c|}{0.4299} & \textbf{0.6943}  \\ \hline
\end{tabular}
\begin{tablenotes}
  \item[*] We noticed a small typo error in our previous version, in which the values for $\rho = \{0.4, 0.6, 0.8\}$ were copied incorrectly. We acknowledge and correct this mistake in this version.
\end{tablenotes}
\end{threeparttable}
}
\label{table:ratio}
\end{table*}

We run first experiment to observe performance of our method compare against DC \cite{data_curation}. Here we select data with rate $\rho\in\{0.2,0.4,0.6,0.8\}$. With each selection rate $\rho$, we define an expected distribution $\mathsf{E}$. The Table \ref{table:ratio} describes the performance of our selection method in detail, with each representative scenario for each selection ratio, as well as the domain-based  and category-based evaluation metrics. The expected distribution varies, and it is entirely determined by human expertise based on application demand. The first purpose of data selection is to select a subset of data that meets an expectation. With DC \cite{data_curation} method, the selection may attempt to select samples with the most diversity and complexity while disregarding the minor appearance category. The results show that our method is better than DC \cite{data_curation} in most categories, selection ratio, and evaluation metrics. 

\begin{figure}[ht]
\centering
    \begin{tikzpicture}
    \begin{axis}[
      xmin=0, xmax=1,
      ymin=0, ymax=1,
      ymajorgrids=true,
      grid style=dashed,
      legend pos=south east,
      width=0.99\linewidth,
      xlabel={Selection ratio $\rho$},
      ylabel={Metric},
    ]

    \addplot[color=green] coordinates {(0.01,0.4906) (0.05,0.6686) (0.1,0.6615) (0.30,0.8038) (0.50,0.8628) (0.67,0.8707) (0.75,0.8958) (0.90,0.9135) (0.95,0.9178) (0.99,0.9337)};
    \addplot[name path=n_top,color=green!10] coordinates {(0.01,0.5118) (0.05,0.711) (0.10,0.6925) (0.30,0.829) (0.50,0.8813) (0.67,0.8842) (0.75,0.9346) (0.90,0.9288) (0.95,0.9673) (0.99,0.948)};
    \addplot[name path=n_down,color=green!10] coordinates {(0.01,0.4513) (0.05,0.5963) (0.10,0.6038) (0.30,0.7228) (0.50,0.846) (0.67,0.847) (0.75,0.8516) (0.90,0.899) (0.95,0.8793) (0.99,0.9223)};
    \addplot[blue!50,fill opacity=0.4] fill between[of=n_top and n_down];

    \addplot[color=red] coordinates {(0.01,0.2066) (0.05,0.356) (0.10,0.3539) (0.30,0.5227) (0.50,0.5268) (0.67,0.5433) (0.75,0.5826) (0.90,0.5975) (0.95,0.6362) (0.99,0.62)};
    \addplot[name path=s_top,color=red!10] coordinates {(0.01,0.2297) (0.05,0.3943) (0.10,0.4185) (0.30,0.5845) (0.50,0.5753) (0.67,0.58) (0.75,0.66) (0.90,0.629) (0.95,0.6785) (0.99,0.6468)};
    \addplot[name path=s_down,color=red!10] coordinates {(0.01,0.093) (0.05,0.3316) (0.1,0.2989) (0.30,0.3946) (0.50,0.501) (0.67,0.493) (0.75,0.522) (0.90,0.59) (0.95,0.5822) (0.99,0.58)};
    \addplot[orange!50,fill opacity=0.4] fill between[of=s_top and s_down];

    \legend{,,,$\mathcal{S}_c$,,,,$\mathcal{S}_d$}
    \end{axis}
    
    \end{tikzpicture}

\caption{Performance of our selection method on selecting data of BDD100K videos in various $\rho$ for some ADAS features.}
\label{fig:curve_spsa}
\end{figure}
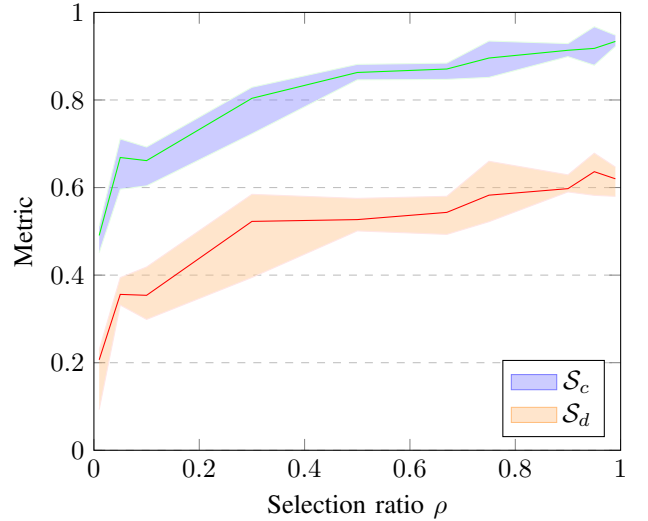

\textbf{Selection for ADAS/ADS validation.}
We select data for validation on some ADAS level 2 and 3 features, such as Lane Departure Warning (LDW), Adaptive Cruise Control (ACC), Autonomous Emergency Braking (AEB), Automated High Beam (AHB), Rear Collision Warning (RCW), Forward Collision Warning (FCW), Lane Changing Assist (LCA), Highway Pilot (HWP), and Intelligent Speed Alerting (ISA). Each of these ADAS features has a set of ODD conditions that require safety validation, as described by NHTSA \cite{nhtsa_fw}, ISO 15622 \cite{iso_15622_2018}, ISO 17361 \cite{iso_17361_2017}, ISO 11270 \cite{iso_11270_2014}, ISO 19237 \cite{iso_19237_2017}, ISO 22839 \cite{iso_22839_2013}. We setup the selected ratio $\rho$ as the following values: $\{0.01, 0.05, 0.1, 0.3, 0.5, 0.67, 0.75, 0.9, 0.95, 0.99\}$. The metadata tag extracted of BDD100K video which are in the list ODD involved of each feature will be kept to assess our method. With each above ADAS feature, we define an expected distribution $\mathsf{E}$ on each selection ratio $\rho$. Finally, we calculate the average score, percentile 70\%, percentile 30\%, and up variance and down variance of $\mathcal{S}_c$ and $\mathcal{S}_d$ as in chart in Figure \ref{fig:curve_spsa} for each selected ratio $\rho$. The figure shows that as $\rho$ increases, the value of both of two metrics increase. This indicates that a low keep ratio makes it more difficult to ensure that the selected data follows the expected distributions. Furthermore, the results show that the variance to score in lower keep ratio $\rho$ increases. It demonstrates that when we aim to select a larger amount of dataset, our selection method is more stable with a variety of expected distributions, including both hard and easy expected distributions. (The hard expected distribution can be considered when there is less data in the entire dataset, but we define the expected ratio to select this data, which is impossible. For example, if we define that the expected ratio of tunnel is 0.5 in the selected dataset with a keep ratio of 0.7, then even if we keep all tunnel data, the actual ratio in the selected data is still much lower than that expected ratio.)

\textbf{Impact of similarity filtering.}
We analyze the impact of similarity filtering on the selection data. We conducted experiments with different values of $\rho \in \{0.1,0.3,0.5,0.7,0.9\}$. In the selection function $\Psi_\rho$, we use both similarity filtering and no filtering. After selecting data, we calculate the mean absolute error (MAE) of each pair of metadata vectors from the selected data. Then we calculate the average MAE of all MAE values for comparison.
The table \ref{tab:sim} shows the average MAE of data selected using our method with and without similarity filtering. The results show that similarity filtering in $\Psi_\rho$ increases the average MAE of all ranges of $\rho$ compared to without filtering. It indicates that the selected data with similarity filtering are more diverse and less similar to each other than without using filtering.
\begin{table}[ht]
\centering
\caption{Average MAE of selected data (w/o: without similarity filtering; w: with similarity filtering)}
\label{tab:sim}
\begin{tabular}{|l|c|c|c|c|c|}
\hline
    & \multicolumn{1}{l|}{$\rho=0.1$} & \multicolumn{1}{l|}{$\rho=0.3$} & \multicolumn{1}{l|}{$\rho=0.5$} & \multicolumn{1}{l|}{$\rho=0.7$} & \multicolumn{1}{l|}{$\rho=0.9$} \\ \hline
w/o & 0.1315                  & 0.1324                  & 0.1632                  & 0.1646                  & 0.188                   \\ \hline
w   & 0.1929                  & 0.2089                  & 0.2197                  & 0.2293                  & 0.2731                 \\ \hline
\end{tabular}
\end{table}

\begin{figure*}[ht]
    \centering
    \begin{subfigure}{.31\textwidth}
        \centering
        \includegraphics[width=.95\linewidth]{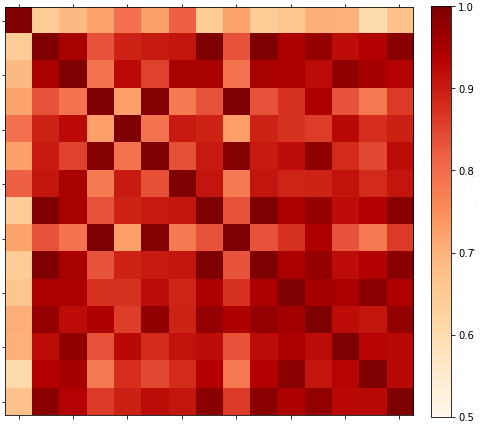}  
        \caption{15 samples of DC \cite{data_curation}}
        \label{subfig:dc}
    \end{subfigure}
    \begin{subfigure}{.31\textwidth}
        \centering
        \includegraphics[width=.95\linewidth]{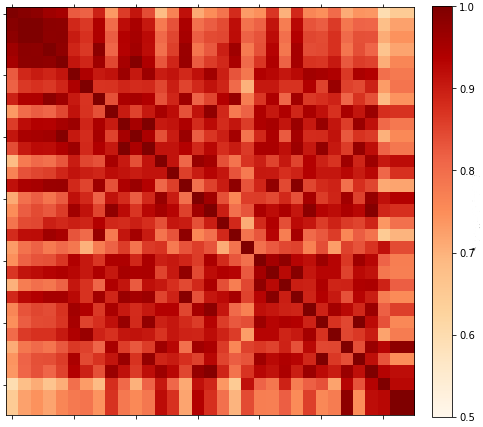}  
        \caption{30 samples of \textbf{Ours.} without filtering}
        \label{subfig:our_without_filter}
    \end{subfigure}
    \begin{subfigure}{.31\textwidth}
        \centering
        \includegraphics[width=.95\linewidth]{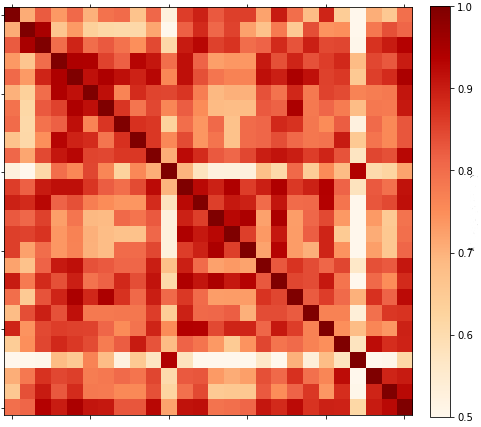}  
        \caption{25 samples of \textbf{Ours.} with filtering}
        \label{subfig:our_with_filter}
    \end{subfigure}
    \caption{The similarity matrix of randomly selected samples in the data selected by DC \cite{data_curation} is compared against our method with or without similarity filtering.}
    \label{fig:sim_analysis}
\end{figure*}

The figure \ref{fig:sim_analysis} shows the qualitative analysis of the impact of similarity filtering in our selected function $\Psi_\rho$. The results show that data selected by DC \cite{data_curation} method are more similar internally than our method. Furthermore, by incorporating similarity filtering in the selected function, our method demonstrates that the data is more diverse and rarely similar to each other than without filtering. It indicates that our method can select data that matches what is expected while also retaining data diversity and complexity. 

\begin{figure*}[ht]
\captionsetup[subfigure]{labelformat=empty}
\begin{minipage}{0.24\linewidth}
    \begin{subfigure}{\linewidth}
        \centering
        \caption{train/0096f810-6bcc27da.mp4}
        \includegraphics[width=\textwidth]{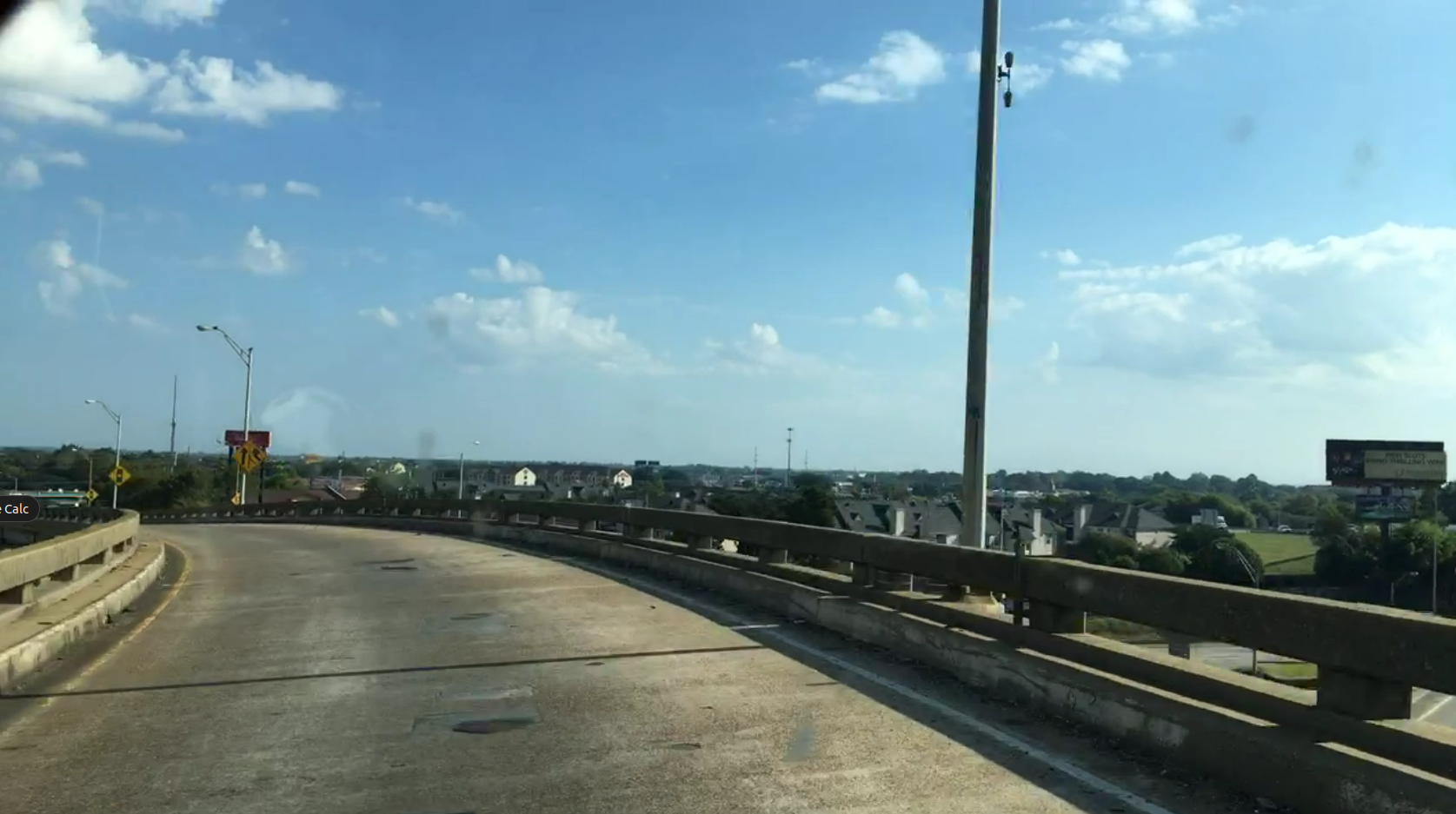}
        
    \end{subfigure}
    \begin{subfigure}{\linewidth}
        \centering
        \caption{Route information on map}
        \includegraphics[width=\textwidth]{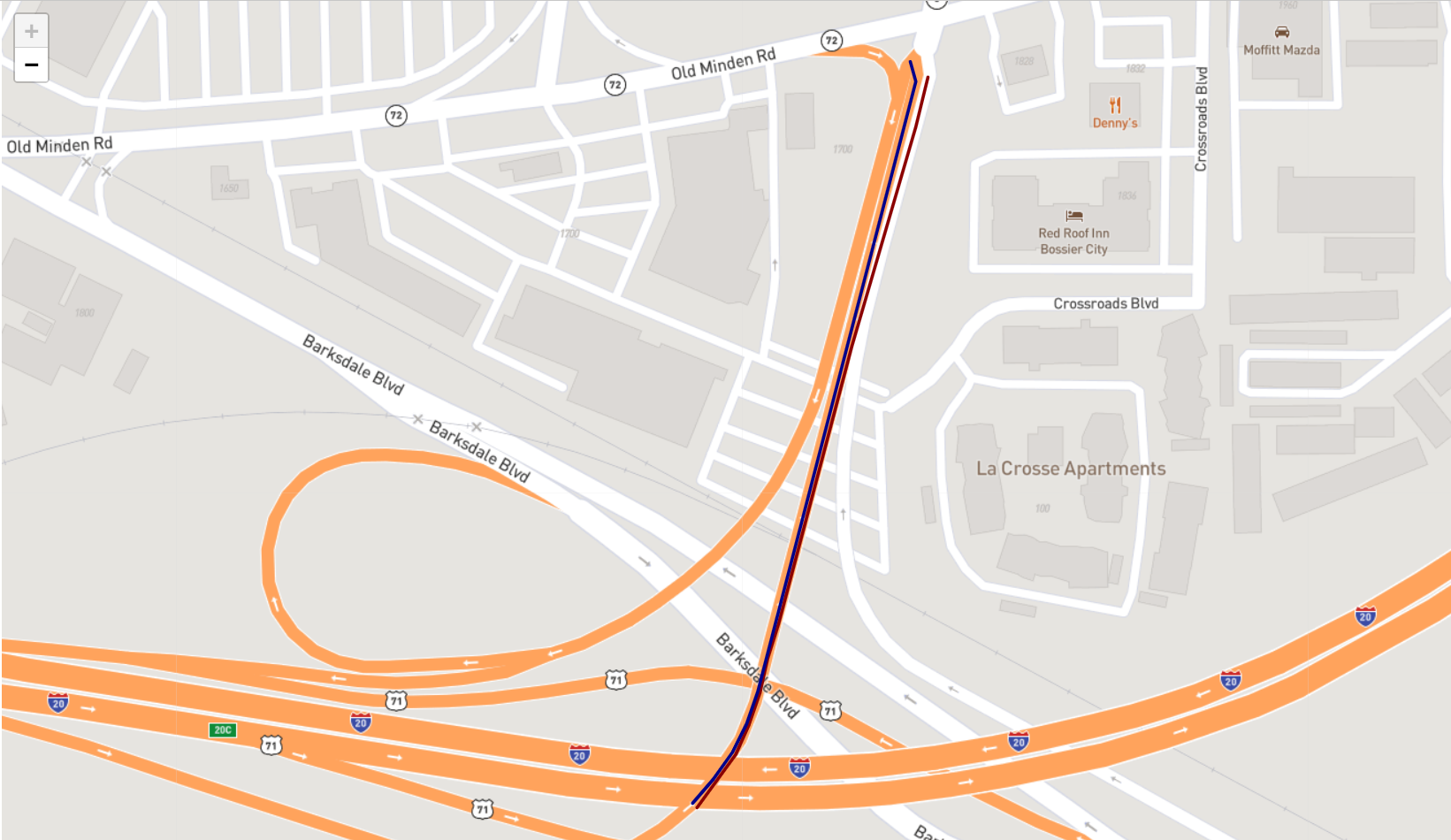}
        
    \end{subfigure}
    \begin{subfigure}{\linewidth}
    \captionsetup{font=tiny}
        \centering
        \caption{Extracted metadata in duration (seconds).}
        \resizebox{\linewidth}{!}{
        
        \begin{tabular}{|c|c|c|cc|}
        \hline
        bridge      & One-way       & toll     & \multicolumn{1}{c|}{tunnel}  & roundabout \\ \hline
        14          & 41            & 0        & \multicolumn{1}{c|}{0}       & 0          \\ \hline
        1-lane      & 2-lanes       & 3-lanes  & \multicolumn{1}{c|}{4-lanes} & highway    \\ \hline
        41          & 0             & 0        & \multicolumn{1}{c|}{0}       & 0          \\ \hline
        Primary-way & Secondary-way & Link-way & \multicolumn{2}{c|}{video duration}                \\ \hline
        0           & 0             & 41       & \multicolumn{2}{c|}{41}                   \\ \hline
        \end{tabular}
        }
    \end{subfigure}

\end{minipage}%
\begin{minipage}{0.24\linewidth}
    \begin{subfigure}{\linewidth}
        \centering
        \caption{train/011ad6b2-f34fd564.mp4}
        \includegraphics[width=\textwidth]{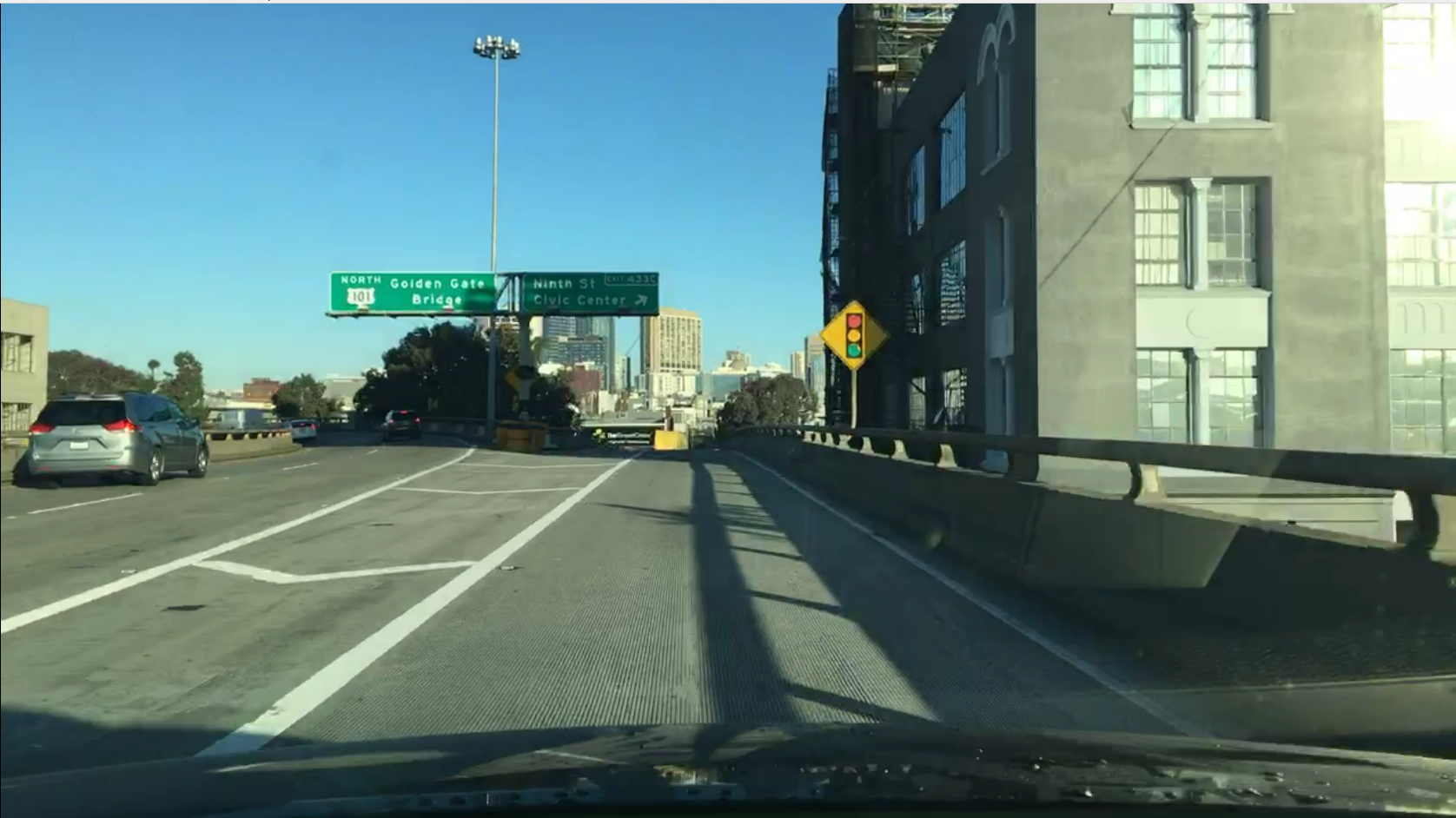}
        
    \end{subfigure}
    \begin{subfigure}{\linewidth}
        \centering
        \caption{Route information on map}
        \includegraphics[width=\textwidth]{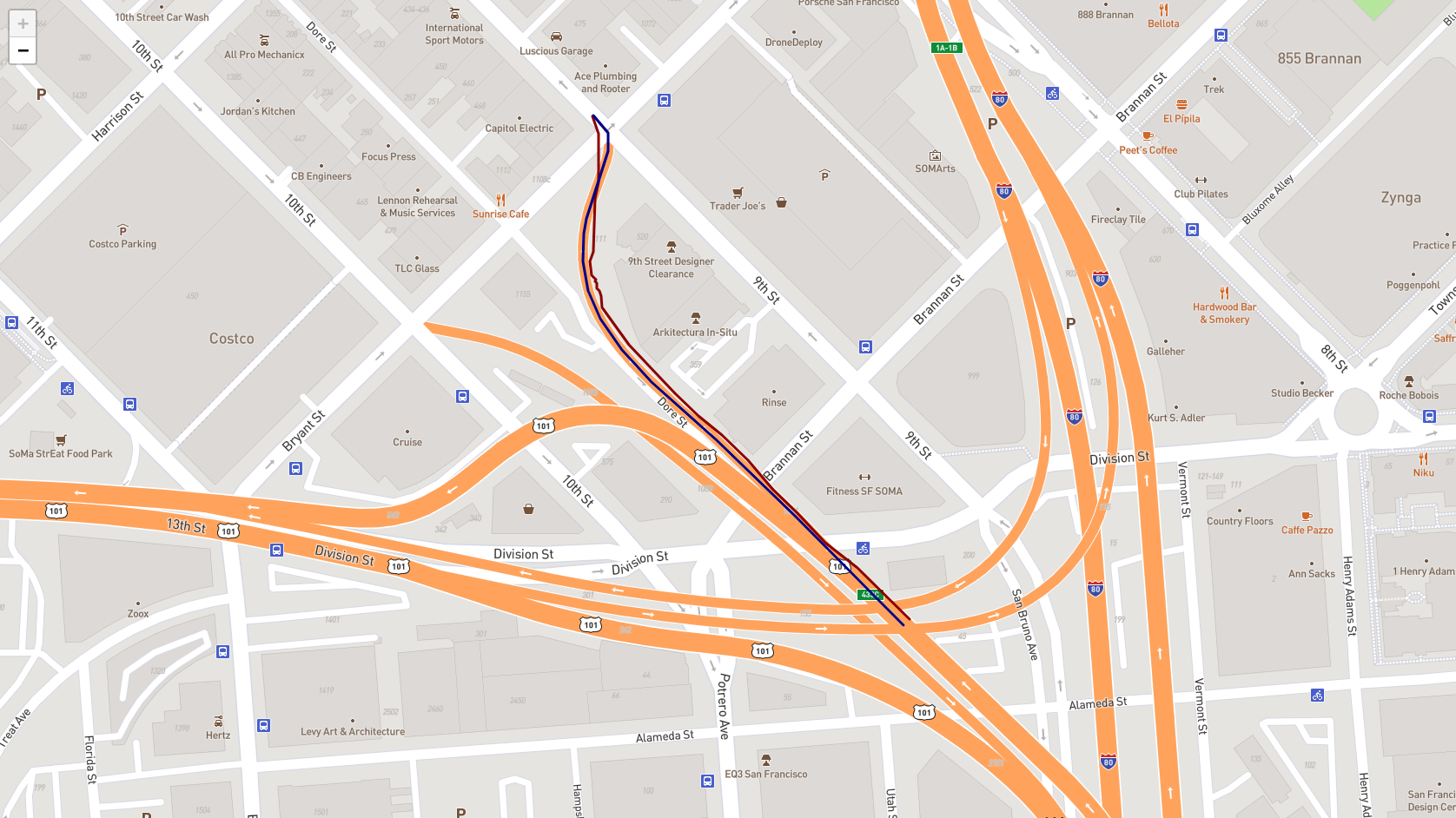}
        
    \end{subfigure}
    \begin{subfigure}{\linewidth}
    \captionsetup{font=tiny}
        \centering
        \caption{Extracted metadata in duration (seconds).}
        \resizebox{\linewidth}{!}{
        
        \begin{tabular}{|c|c|c|cc|}
        \hline
        bridge      & One-way       & toll     & \multicolumn{1}{c|}{tunnel}  & roundabout \\ \hline
        32          & 41            & 0        & \multicolumn{1}{c|}{0}       & 0          \\ \hline
        1-lane      & 2-lanes       & 3-lanes  & \multicolumn{1}{c|}{4-lanes} & highway    \\ \hline
        7           & 28            & 0        & \multicolumn{1}{c|}{6}       & 2          \\ \hline
        Primary-way & Secondary-way & Link-way & \multicolumn{2}{c|}{video duration}                \\ \hline
        0           & 1             & 38       & \multicolumn{2}{c|}{41}                   \\ \hline
        \end{tabular}
        }
    \end{subfigure}
    
\end{minipage}%
\begin{minipage}{0.24\linewidth}
    \begin{subfigure}{\linewidth}
        \centering
        \caption{train/018f06cd-834da3b3.mp4}
        \includegraphics[width=\textwidth]{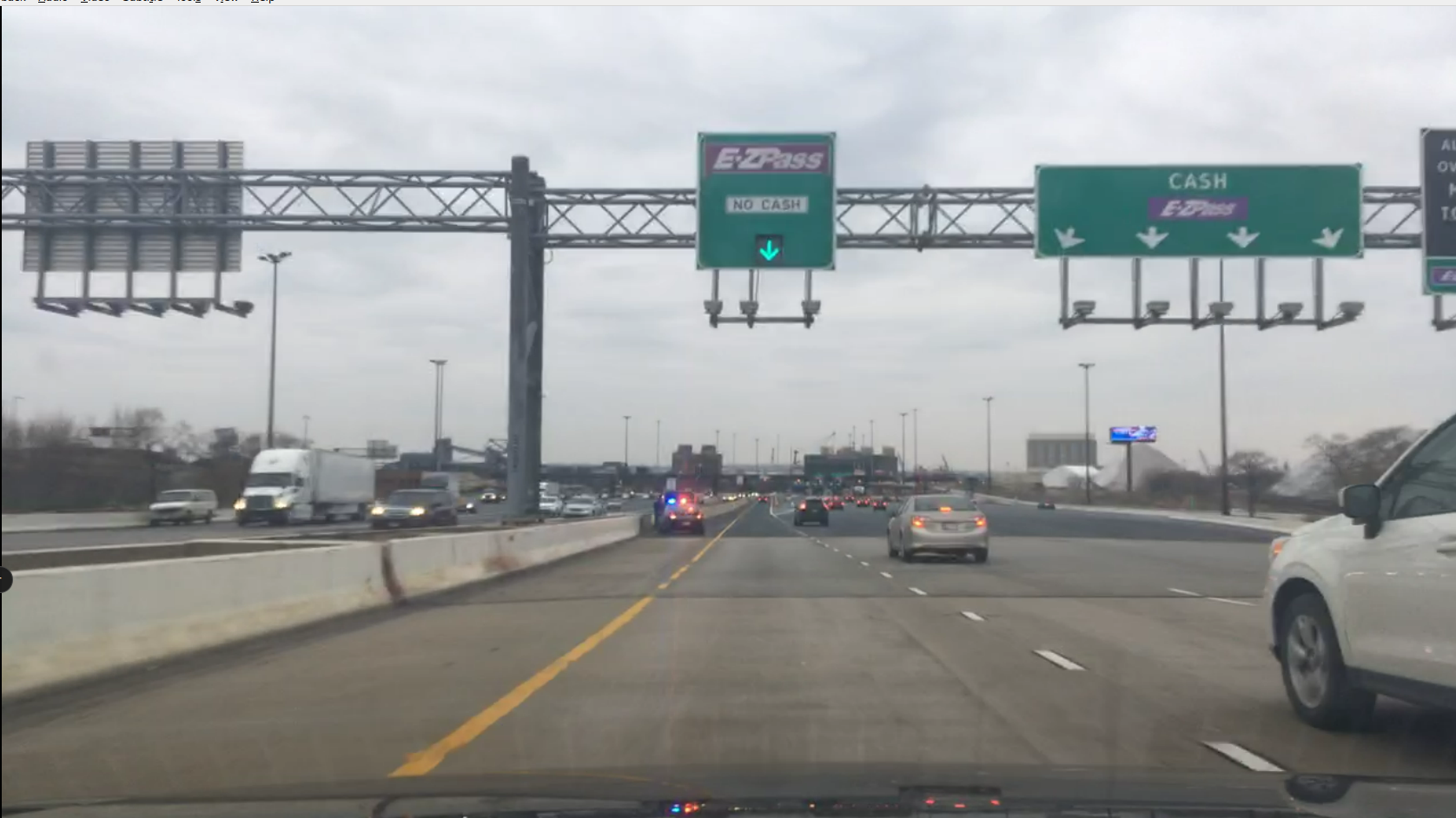}
        
    \end{subfigure}
    \begin{subfigure}{\linewidth}
        \centering
        \caption{Route information on map}
        \includegraphics[width=\textwidth]{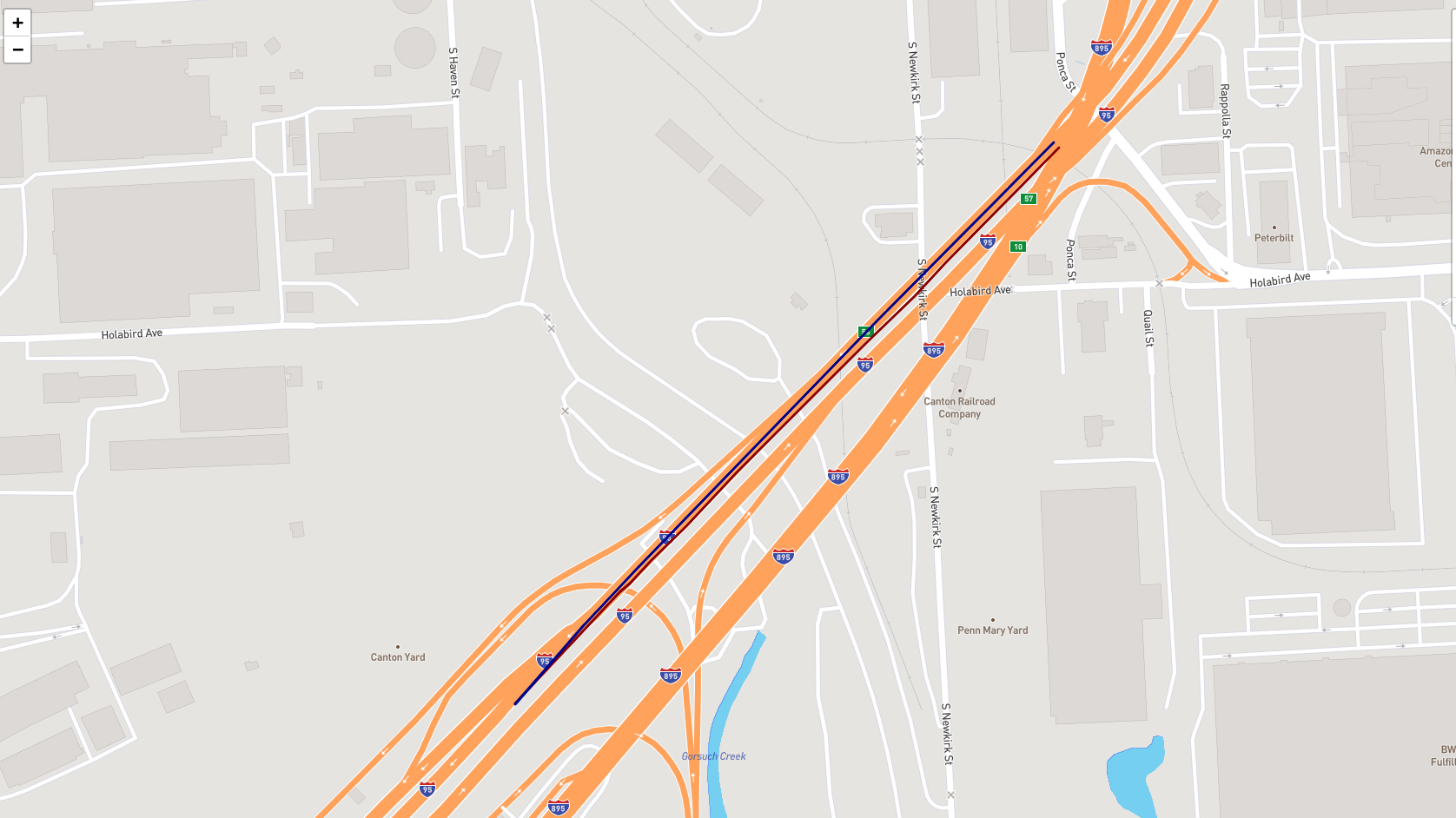}
        
    \end{subfigure}
    \begin{subfigure}{\linewidth}
    \captionsetup{font=tiny}
        \centering
        \caption{Extracted metadata in duration (seconds).}
        \resizebox{\linewidth}{!}{
        
        \begin{tabular}{|c|c|c|cc|}
        \hline
        bridge      & One-way       & toll     & \multicolumn{1}{c|}{tunnel}  & roundabout \\ \hline
        34          & 34            & 29       & \multicolumn{1}{c|}{0}       & 0          \\ \hline
        1-lane      & 2-lanes       & 3-lanes  & \multicolumn{1}{c|}{4-lanes} & highway    \\ \hline
        7           & 0             & 0        & \multicolumn{1}{c|}{24}      & 41         \\ \hline
        Primary-way & Secondary-way & Link-way & \multicolumn{2}{c|}{video duration}       \\ \hline
        0           & 0             & 0        & \multicolumn{2}{c|}{41}                   \\ \hline
        \end{tabular}
        }
    \end{subfigure}
    
\end{minipage}%
\begin{minipage}{0.24\linewidth}
    \begin{subfigure}{\linewidth}
        \centering
        \caption{train/01079b7f-9d0e53c5.mp4}
        \includegraphics[width=\textwidth]{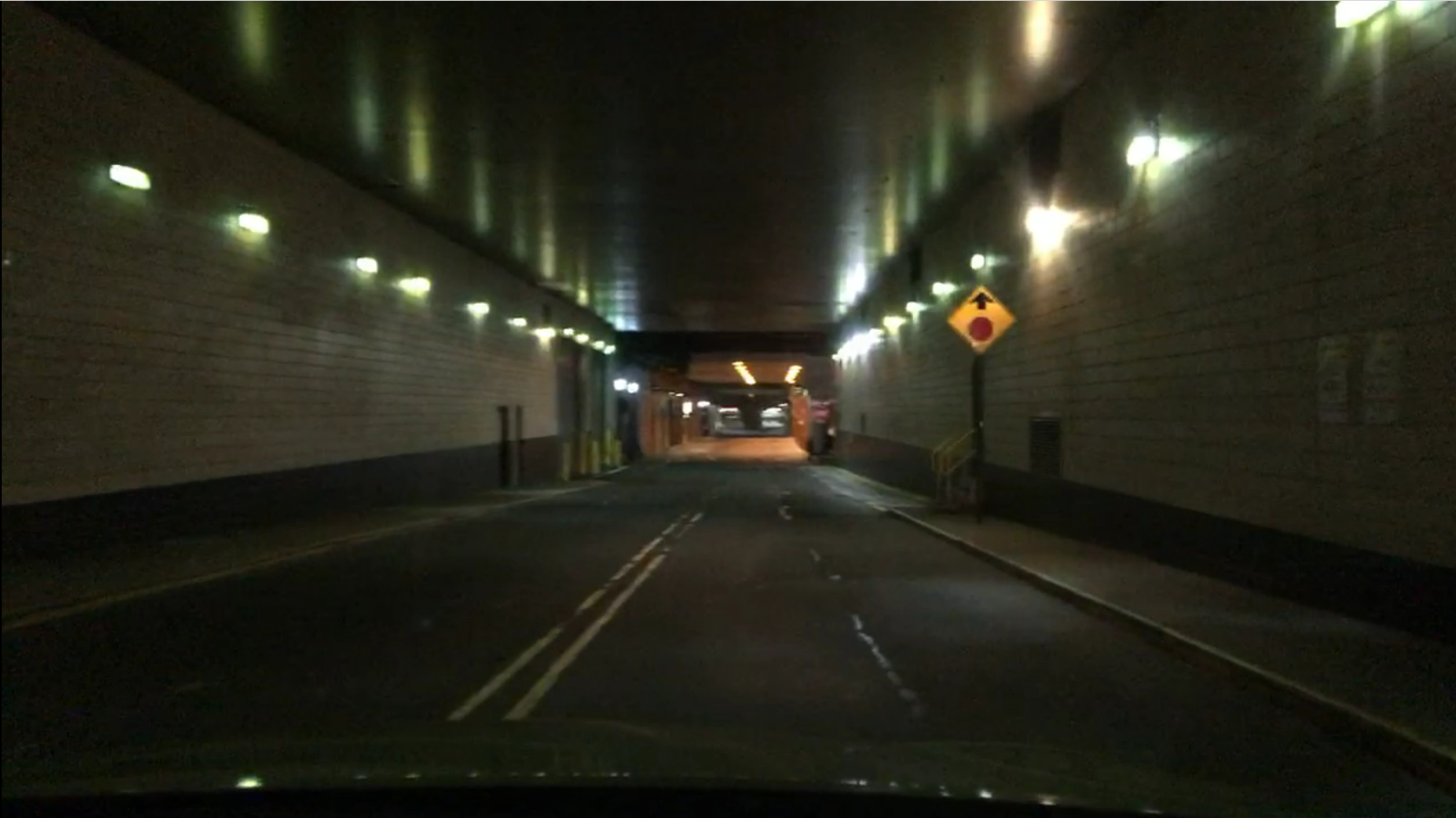}
        
    \end{subfigure}
    \begin{subfigure}{\linewidth}
        \centering
        \caption{Route information on map}
        \includegraphics[width=\textwidth]{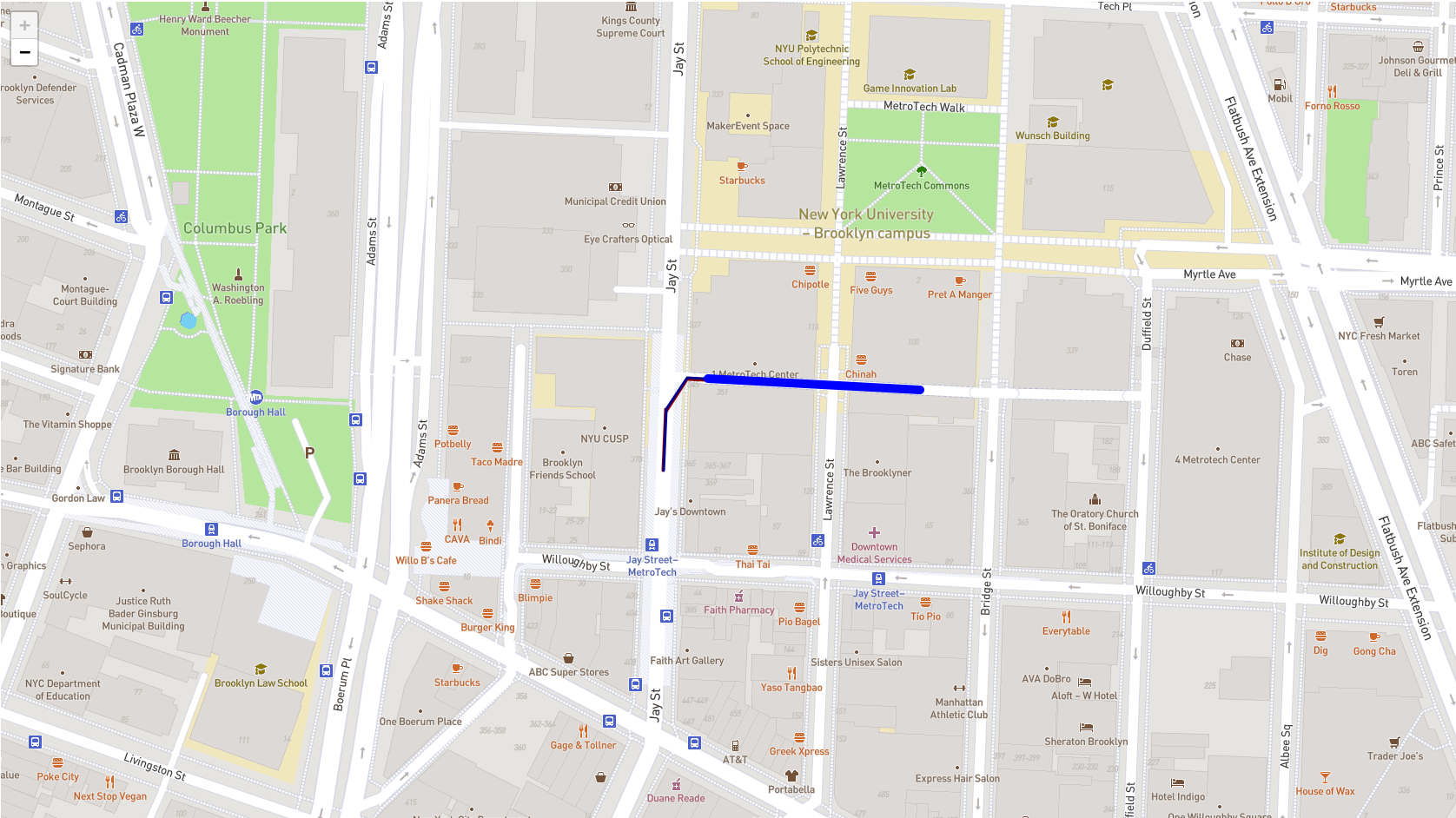}
        
    \end{subfigure}
    \begin{subfigure}{\linewidth}
    \captionsetup{font=tiny}
        \centering
        \caption{Extracted metadata in duration (seconds).}
        \resizebox{\linewidth}{!}{
        
        \begin{tabular}{|c|c|c|cc|}
        \hline
        bridge      & One-way       & toll     & \multicolumn{1}{c|}{tunnel}  & roundabout \\ \hline
        0           & 0             & 0        & \multicolumn{1}{c|}{23}      & 0          \\ \hline
        1-lane      & 2-lanes       & 3-lanes  & \multicolumn{1}{c|}{4-lanes} & highway    \\ \hline
        41          & 0             & 0        & \multicolumn{1}{c|}{0}       & 0          \\ \hline
        Primary-way & Secondary-way & Link-way & \multicolumn{2}{c|}{video duration}       \\ \hline
        0           & 0             & 0        & \multicolumn{2}{c|}{41}                   \\ \hline
        \end{tabular}
        }
    \end{subfigure}
    
\end{minipage}%
\caption{Qualitative example of selected data on BDD100K video by our selection method with OpenStreetMap.}
\label{fig:qualitative_bdd_map}
\end{figure*}

\textbf{Qualitative analysis.}
We randomly select a sample of data from the BDD100K videos dataset using our method for qualitative analysis. The detail of these examples are shown in Figure \ref{fig:qualitative_bdd_map}. With each video, we use the GPS information provided by this dataset and OpenStreetMap, an open source map, to query the road attributes associated with the video. The driving route of data collection in each video is shown on the map in this Figure using GPS. We query the metadata set using OpenStreetMap: bridge, one-way, toll, tunnel, roundabout, 1/2/3/4-lanes, highway/primary/secondary/link-ways. We summarize the duration of each metadata in seconds, along with the video duration, after extracting these metadata information for entire GPS data points of video. Finally, the metadata duration information is used in our selection method to calculate the score of each video sample based on the optimal model in order to rank videos and make further selections. It demonstrate that our method can extract complexity and diversity data from the original dataset.

\section{CONCLUSIONS}\label{sec:conclude}
In this paper, we present our proposed data selection method for validation of variety of ADAS/ADS features in autonomous vehicles. We introduced our framework for data selection based on the metadata of data set. Furthermore, we proposed an algorithm to train a model which is used to optimize the similarity of metadata distribution of selected data and a predefined metadata distribution that is expected for a validation task. Additionally, we provide two metrics which can be used to evaluate the quality of data selection, and guide the data selection algorithm. In experiment, we used a large self-driving dataset BDD100K which consists of near thousands hours driving. Experiment results of data selection compared with other methods indicates that our method is efficient, highly reliable and can be used for validation of a variety of self-driving safety functions. 



\section*{ACKNOWLEDGMENT}
This work was realized in imec.ICON Hybrid AI for predictive road maintenance (HAIROAD) project, with the financial support of Flanders Innovation \& Entrepreneurship (VLAIO, project no. HBC.2023.0170).


\bibliographystyle{IEEEtran}
\bibliography{reference}

\end{document}